\documentclass[sigconf]{acmart}

\pdfoutput=1

\usepackage{booktabs} % For formal tables

\usepackage{wrapfig}
\usepackage{ifthen}

\setcopyright{rightsretained}
\acmConference[DEBS'18]{12th ACM International Conference on Distributed and Event-Based Systems}{June 2018}{Hamilton, New Zealand}
\acmYear{2018}
\copyrightyear{2018}

\newboolean{showcomments}
\setboolean{showcomments}{true}
\ifthenelse{\boolean{showcomments}}
{ \newcommand{\mynote}[3]{
    \fbox{\bfseries\sffamily\scriptsize#1}
    {\small$\blacktriangleright$\textsf{\emph{\color{#3}{#2}}}$\blacktriangleleft$}}}
{ \newcommand{\mynote}[3]{}}
\begin{document}
\title{Grand Challenge: Predicting Destinations by Nearest Neighbor Search on Training Vessel Routes}
%\titlenote{Produces the permission block, and copyright information}
%\subtitle{Extended Abstract}
%\subtitlenote{The full version of the author's guide is available as \texttt{acmart.pdf} document}
\renewcommand{\shorttitle}{Predicting Destinations by Nearest Neighbor Search on Training Vessel Routes}

\author{Valentin Ro\c{s}ca}
\affiliation{%
  \institution{Alexandru Ioan Cuza University of Ia\c{s}i, Romania}
}
\email{valentin.rosca@info.uaic.ro}

\author{Emanuel Onica}
\affiliation{%
  \institution{Alexandru Ioan Cuza University of Ia\c{s}i, Romania}
}
\email{eonica@info.uaic.ro}

\author{Paul Diac}
\affiliation{%
  \institution{Alexandru Ioan Cuza University of Ia\c{s}i, Romania}
}
\email{paul.diac@info.uaic.ro}

\author{Ciprian Amariei}
\affiliation{%
  \institution{Alexandru Ioan Cuza University of Ia\c{s}i, Romania}
}
\email{ciprian.amariei@gmail.com}

\begin{abstract}
The DEBS Grand Challenge 2018 is set in the context of maritime route prediction. Vessel routes are modeled as streams of Automatic Identification System (AIS) data points selected from real-world tracking data. The challenge requires to correctly estimate the destination ports and arrival times of vessel trips, as early as possible. Our proposed solution partitions the training vessel routes by reported destination port and uses a nearest neighbor search to find the training routes that are closer to the query AIS point. %The predicted port is the reported port of the closest training route found. The response arrival time is the current time of the query AIS point plus the remaining time spent on the closest route found until its final arrival%
Particular improvements have been included as well, such as a way to avoid changing the predicted ports frequently within one query route and automating the parameters tuning by the use of a genetic algorithm. This leads to significant improvements on the final score.

\end{abstract}

\begin{CCSXML}
<ccs2012>
<concept>
<concept_id>10010147.10010257.10010321</concept_id>
<concept_desc>Computing methodologies~Machine learning algorithms</concept_desc>
<concept_significance>300</concept_significance>
</concept>
<concept>
<concept_id>10002951.10003227.10003236</concept_id>
<concept_desc>Information systems~Spatial-temporal systems</concept_desc>
<concept_significance>300</concept_significance>
</concept>
<concept>
<concept_id>10010405.10010481.10010485</concept_id>
<concept_desc>Applied computing~Transportation</concept_desc>
<concept_significance>300</concept_significance>
</concept>
<concept>
<concept_id>10010147.10010257.10010293.10011809.10011812</concept_id>
<concept_desc>Computing methodologies~Genetic algorithms</concept_desc>
<concept_significance>300</concept_significance>
</concept>
<concept>
<concept_id>10010147.10010178.10010205.10010207</concept_id>
<concept_desc>Computing methodologies~Discrete space search</concept_desc>
<concept_significance>300</concept_significance>
</concept>
</ccs2012>
\end{CCSXML}

\ccsdesc[300]{Computing methodologies~Machine learning algorithms}
\ccsdesc[300]{Information systems~Spatial-temporal systems}
\ccsdesc[300]{Applied computing~Transportation}
\ccsdesc[300]{Computing methodologies~Genetic algorithms}
\ccsdesc[300]{Computing methodologies~Discrete space search}

\keywords{machine learning, ball trees, genetic algorithm, space partitioning, nearest neighbor}

\copyrightyear{2018} 
\acmYear{2018} 
\setcopyright{acmcopyright}
\acmConference[DEBS '18]{The 12th ACM International Conference on Distributed and Event-based Systems}{June 25--29, 2018}{Hamilton, New Zealand}
\acmBooktitle{DEBS '18: The 12th ACM International Conference on Distributed and Event-based Systems, June 25--29, 2018, Hamilton, New Zealand}
\acmPrice{15.00}
\acmDOI{10.1145/3210284.3220509}
\acmISBN{978-1-4503-5782-1/18/06}

%\CopyrightYear{2018} 
%\setcopyright{acmcopyright} 
%\conferenceinfo{DEBS '18,}{June 25--29, 2018, Hamilton, New Zealand}
%\isbn{978-1-4503-5782-1/18/06}\acmPrice{$15.00}
%\doi{https://doi.org/10.1145/3210284.3220509}

\maketitle

\section{Introduction}
\label{sec:intro}

%The Grand Challenge is part of the DEBS conference since 2011. 
The 2018 edition of DEBS Grand Challenge targets a route prediction problem, using real-world data published by the maritime transportation service Marine Traffic. 
%Geographically, all routes and ports are within the Mediterranean Sea area. 
Part of the routes are used for training. For these the arrival ports and times are known in advance. The information is provided as Automatic Identification System (AIS) data. AIS points include geographic coordinates, ship id, current speed, timestamp, ship type, ship direction (course and heading) and draught. For each evaluated AIS point the response is split into estimating the arrival port (Query 1) and time (Query 2). % that are evaluated in separate runs.
More information about the problem proposed is available in~\cite{gulisano2018debsgc}.

Our solution stores paths of ships from training routes separately for each destination port. All AIS points are added in a Spatial Indexing Structure optimized for fast retrieval of nearest point and insertion of new points. This structure is used for both queries: finding the nearest point to a query point will provide a route for which the destination port and arrival time are known.
\vspace{-3pt}
\section{Architecture overview}
\label{sec:architecture}

The solution is split into training and classification phases.
Both expose a series of configurable parameters and optional features (e.g., switching from one metric to another, usage of different dimensions to represent space). Finding the best combination gets harder as the parameters number increase. To ease this process we used a genetic algorithm in which individuals are assignments of parameters and the fitness function is the overall prediction score. We ran the genetic algorithm locally on training data and observed that the score increased, therefore, validating this approach of setting the optimal values for parameters.

\vspace{-3pt}
\subsection{Training Phase}

\subsubsection{Route Partitioning}
First we divide the data into routes. A route is a set of AIS points emitted by a single ship when moving from one port to another. In the context of our training data, we identified the routes as a series of AIS points having the same ship id, departure port and arrival time. Then we sort the data points in each route by reported timestamp and construct a linked list. We consider each list node as a \emph{RoutePoint} object that has a unique id and a precomputed distance from departure, i.e. the sum of the great circle distances between all pairs of previous consecutive points that are part of the same route.

\subsubsection{Port Partitioning and 5-Dimensional Space}
\label{sec:fivedimensionalspace}
We partition all the \emph{RoutePoint}s by their arrival port. For each port we construct a separate Ball Tree \cite{nielsen2009tailored} that contains all \emph{RoutePoint}s of routes that arrive to it. All \emph{RoutePoint} objects contain references to the current and previous AIS point in the route. Using these we calculate the \emph{bearing} that is the polar angle between current and previous points.

Ball Trees can work with any number of dimensions. The best choice found empirically was to use five dimensions. First three dimensions are mapping \emph{(latitude, longitude)} to \emph{(x,y,z)} in a 3-D cartesian space centered in the Earth center. The last two dimensions are the \emph{sine} and \emph{cosine} values of the \emph{bearing} angle. Each dimension is graded with a \emph{magnitude} parameter that is a floating point in \emph{[0, 1]}. These are some of the values optimized by the genetic algorithm.

\vspace{-3pt}
\subsection{Classification Phase}

\begin{figure}[h]
\includegraphics[width=0.40\textwidth]{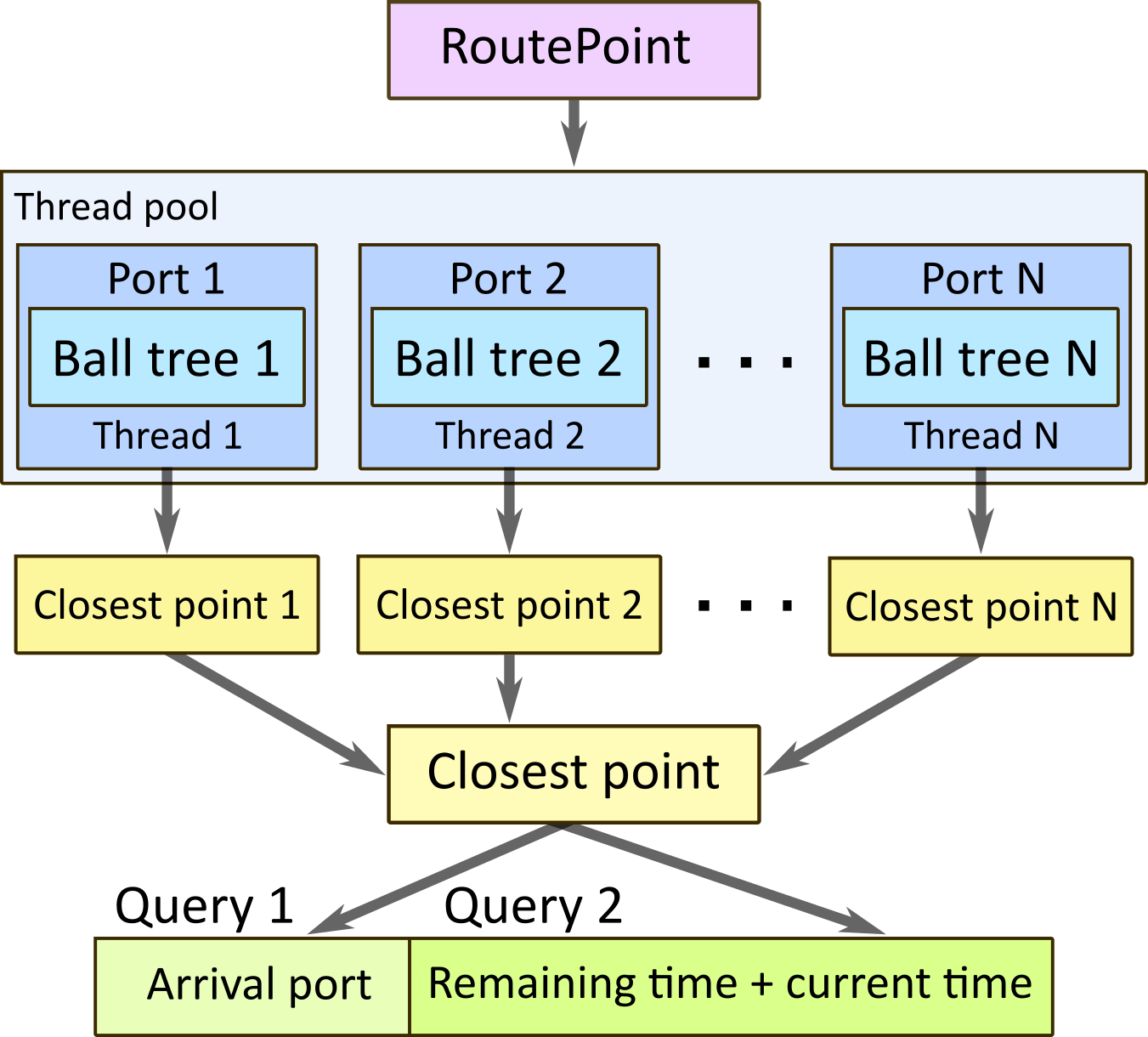}
\caption{Evaluation of a route point: get the closest point first for each possible destination and then the closest of all these. Use this point as reference to respond to queries.}
\label{fig:arch}
\vspace{-10pt}
\end{figure}

\subsubsection{Nearest RoutePoints search}
A high level view of our algorithm is depicted in Figure~\ref{fig:arch}.
Given the \emph{RoutePoint} to classify, based on the 5-dimensional Euclidean metric, we search for the closest \emph{RoutePoint} in each of the Ball trees. This is parallelized using Java Parallel Streams~\cite{urma2014java}. Then we use a \emph{similarity function} to get the closest point out of all these.

The \emph{similarity function} has as input two \emph{RoutePoints} %: the one to classify and the corresponding closest associated to a port, 
and returns a similarity factor.
This factor is based on the distance between points, but also takes into account the course, heading, speed and distance from departure. The function sets a weight given by each of these parameters based on a penalty factor chosen by a genetic algorithm and adjusted proportionally with the difference between the corresponding parameter values in the two points (e.g., the weight adjustment will make the course more significant as closer the course of one point will get to the course of the other point). The multiplication of the parameter weights is used as an aggregated weight on the great circle distance between the two points. This helps to distinguish which is the more relevant result in case of relatively similar distances to points associated to different ports. % The smaller the result, the closer the points are to each other.

Using the \emph{similarity function}, we chose the final \emph{RoutePoint} that best represents the given \emph{RoutePoint} to classify. For Query 1, the result is the arrival port of the chosen \emph{RoutePoint}. For Query 2 we use the precomputed remaining time spent on the route from the chosen \emph{RoutePoint} to the route end. The time is added to the timestamp of the \emph{RoutePoint} to classify.

\subsubsection{Longest Predicted Segment}
Because the final score is based on the longest correct suffix, GPS errors and variance in AIS points emission intervals between routes can lower the score drastically. As such, we remember the prediction of the classifier and actually output the value of the longest consecutive subsequence with the same prediction. This reduces the fluctuations of the prediction while still allowing the change of the chosen results when the prediction becomes stable.

%\subsection{Spatial Indexing Structure (SIS)}
%split per port, describe all details here, what are ball trees and why we used them

%\subsection{Sliding window}
%we avoid changing the estimated arrival port too much because of the correct prefix evaluation scheme

%\subsection{Query 1 solution}
%Figure 1: architecture overview
%\subsubsection{Search on SIS}
%\subsubsection{Similarity metrics}
%(lat, lon) become (x,y,z) in 3-D space, and bearing is split to sin, cos. Finally the ball trees are 5-dimensional and each dimension is also multiplied with a parameter.

%\subsection{Query 2 solution}
%Figure 2: Message Eval Flow

%Hmmm actually here the solution is simple but apperently %efficient. I think no figure will be required but we will see.

%\subsection{Multithreading usage}
%this will be short but relevant

\vspace{-3pt}
\section{Evaluation}
\label{sec:Evaluation}

For the implementation of the Spatial Indexing Structure we tested two different data structures: first K-D trees and then Ball trees. Switching to Ball trees reduced the runtime by 41.1\% (average processing time of one evaluation route on our local benchmark).

By adding the two dimensions that represent the \emph{bearing} measure: sine and cosine of the angle; the score of Query 1 was improved by 4.5\%, while the Query 2 score declined by 19.6\%, relative to the use of \emph{bearing} as a single dimension.

Parallelizing the computation of the nearest point by multi-threading reduced the run time by 40.4\% (average processing time of one evaluation route).

Our best scores obtained on the testing platform are \textbf{0.8249} average earliness rate for Query 1 with \textbf{19.0} seconds total system runtime, and \textbf{181.866} mean error in minutes for Query 2 with \textbf{18.0} seconds total system runtime.

%We tested our solution with several changes on the Hobbit platform and kept the changes that improved the score on the benchmark. First we used KD-Trees for the Spatial Indexing Structure, and later tried Ball Trees that produced faster running times. Also, changing the dimensional space of Ball Trees to the 5-dimension metric described in \ref{BallTreeLabel} produced an increase in the Q1 score of 30\%.

%Some interesting metrics would be: score with and without 5-dimensional Ball trees, longest-segment-estimated trick, running time with and without multithreading. Better compare relative, e.g. like x% decrease in running time, etc.
\vspace{-3pt}
\section{Future Work}
\label{sec:future}

There are a series of potential improvements to our solution. One is the use of R-Trees as the Spatial Indexing Structure as they are also optimized for nearest neighbor search but also especially for the great circle distance.

A higher level improvement is dynamic learning of evaluation routes. %The end of any evaluation route can be determined by checking the entrance within the bounding box of a port. So at this time 
When the destination port and arrival time are known, the route can be inserted in the Spatial Indexing Structure similarly to a training route.

Lastly, instead of looking only at the most recent AIS point of a route, we could consider a sliding window of the last \emph{k} AIS points. For this, the distance function can be generalized to sum the distances within last \emph{k} most recent pairs of points.  However implementing this would require adapting the \emph{similarity function} and the run time can potentially increase.

\section*{Acknowledgements}
\setlength{\columnsep}{9pt}
\setlength{\intextsep}{3pt}
\begin{wrapfigure}{r}{0.09\textwidth}
\includegraphics[width=\linewidth]{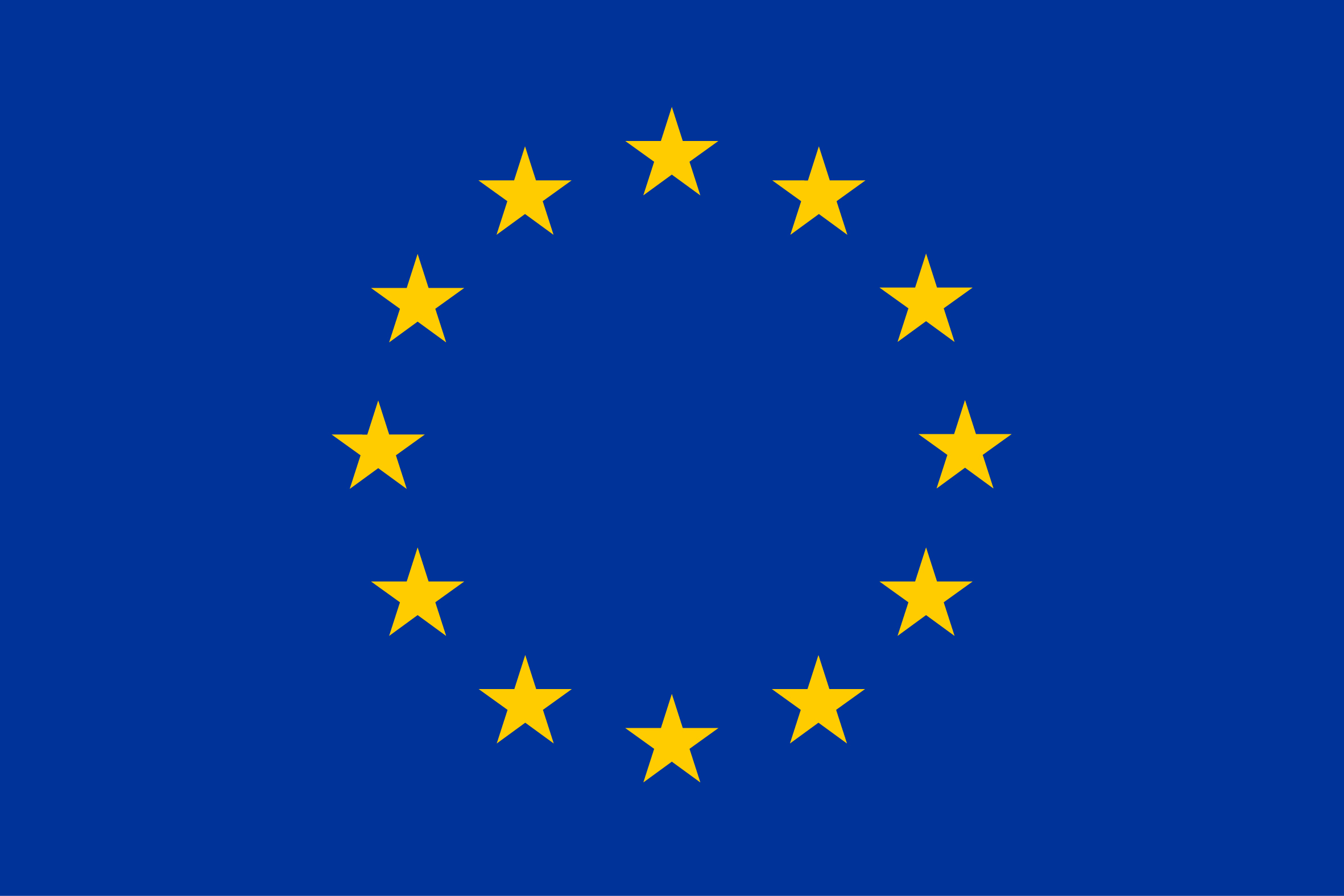}
\end{wrapfigure}
The dissemination of this work is partly funded by the \emph{European Union{\textquotesingle}s Horizon 2020 research and innovation programme} under grant agreement No 692178. This work was also partly supported by a grant of the Romanian National Authority for Scientific Research and Innovation, CNCS/CCCDI - UEFISCDI, project number 10/2016, within PNCDI III.

%TODO: use the references.bib file for references (currently does not work for some reason).
\vspace{-3pt}
\bibliographystyle{ACM-Reference-Format}
\bibliography{references}

\end{document}